\pdfoutput=1

\documentclass[11pt]{article}

\usepackage{arabtex}
\usepackage{utf8}

\usepackage[]{ACL2023}

\usepackage{times}
\usepackage{latexsym}

\usepackage{booktabs}
\usepackage{arydshln}
\usepackage{amsmath}
\usepackage{graphicx}
\usepackage{mathabx}
\usepackage{setspace}
\usepackage{xcolor}
\usepackage{stmaryrd}
\usepackage{mathtools} %

\newcommand{\veryshortarrow}[1][3pt]{\mathrel{%
   \hbox{\rule[\dimexpr\fontdimen22\textfont2-.2pt\relax]{#1}{.4pt}}%
   \mkern-4mu\hbox{\usefont{U}{lasy}{m}{n}\symbol{41}}}}

\usepackage[utf8]{inputenc}
\usepackage[main=english, arabic, russian]{babel}
\usepackage{CJKutf8}

\usepackage[T1]{fontenc}

\usepackage[utf8]{inputenc} %

\usepackage{microtype}

\usepackage{inconsolata}

\title{Boosting Zero-shot Cross-lingual Retrieval by Training on\\Artificially Code-Switched Data}

\author{Robert Litschko ~~ Ekaterina Artemova ~~ Barbara Plank \\
        MaiNLP, Center for Information and Language Processing (CIS), LMU Munich, Germany \\ 
        \texttt{\{robert.litschko, ekaterina.artemova, b.plank\}@lmu.de}
        }

\begin{document}
\maketitle
\begin{abstract}
Transferring information retrieval (IR) models from a high-resource language (typically English) to other languages in a zero-shot fashion has become a widely adopted approach. In this work, we show that the effectiveness of zero-shot rankers diminishes when queries and documents are present in different languages. Motivated by this, we propose to train ranking models on artificially code-switched data instead, which we generate by utilizing bilingual lexicons. To this end, we experiment with lexicons induced from (1) cross-lingual word embeddings and (2) parallel Wikipedia page titles. We use the mMARCO dataset to extensively evaluate reranking models on 36 language pairs spanning Monolingual IR (MoIR), Cross-lingual IR (CLIR), and Multilingual IR (MLIR). 
Our results show that code-switching can yield consistent and substantial gains of 5.1 MRR@10 in CLIR and 3.9 MRR@10 in MLIR, while maintaining stable performance in MoIR. 
Encouragingly, the gains are especially pronounced for distant languages (up to 2x absolute gain). We further show that our approach is robust towards the ratio of code-switched tokens and also extends to unseen languages. Our results demonstrate that training on code-switched data is a cheap and effective way of generalizing zero-shot rankers for cross-lingual and multilingual retrieval. 
\end{abstract}

\section{Introduction}
\label{s:intro}
Cross-lingual Information Retrieval (CLIR) is the task of retrieving relevant documents written in a language different from a query language. The large number of languages and limited amounts of training data pose a serious challenge for training ranking models. Previous work address this issue by using machine translation (MT), effectively casting CLIR into a noisy variant of monolingual retrieval \citep{li-cheng-2018-learning,shi-etal-2020-cross,shi-etal-2021-cross,cross-lingual2021}. MT systems are used to either train ranking models on translated training data (\textit{translate train}), or by translating queries into the document language at retrieval time (\textit{translate test}). However, CLIR approaches relying on MT systems are limited by their language coverage. Because training MT models is bounded by the availability of parallel data, it does not scale well to a large number of languages. Furthermore, using MT for IR has been shown to be prone to propagation of unwanted translation artifacts such as topic shifts, repetition, hallucinations and lexical ambiguity \citep{artetxe-etal-2020-translation,litschko-etal-2022-parameter,li-etal-2022-museclir}. In this work, we propose a resource-lean MT alternative to bridge the language gap and propose to use  \textit{artificially code-switched} data.

We focus on zero-shot cross-encoder (CE) models for reranking~\cite{macavaney-etal-2020teaching,jiang-etal-2020-cross}. 
Our study is motivated by the observation that the performance of CEs diminishes when they are transferred into CLIR and MLIR as opposed to MoIR. We hypothesize that training on queries and documents from the same language leads to \textit{monolingual overfitting} where the ranker learns features, such as exact keyword matches, which are useful in MoIR but do not transfer well to CLIR and MLIR setups due to the lack of lexical overlap \citep{litschko2022cross}. In fact, as shown by \citet{roy-etal-2020-lareqa} on bi-encoders, representations from zero-shot models are weakly aligned between languages, where models prefer non-relevant documents in the same language over relevant documents in a different language. To address this problem, we propose to use code-switching as an inductive bias to regularize monolingual overfitting in CEs.

Generation of synthetic code-switched data has served as a way to augment data in cross-lingual setups in a number of NLP tasks~\cite{singh2019xlda,einolghozati-etal-2021-el,tan-joty-2021-code}. They utilize substitution techniques ranging from simplistic re-writing in the target script \cite{gautam-etal-2021-comet}, looking up bilingual lexicons \cite{tan-joty-2021-code} to MT \cite{tarunesh-etal-2021-machine}. 
Previous work on improving zero-shot transfer for IR includes weak supervision \citep{shi-etal-2021-cross}, tuning the pivot language \citep{turc2021revisiting}, multilingual query expansion \citep{blloshmi-etal-2021-ir} and cross-lingual pre-training~\citep{Yang_Ma_Zhang_Wu_Li_Zhou_2020,yu-etal-2021-clir-pretraining,yang-etal-2022-c3,lee2023c}. 
To this end, code-switching is complementary to existing approaches. Our work is most similar to \citet{shi-etal-2020-cross}, who use bilingual lexicons for full term-by-term translation to improve MoIR. 
Concurrent to our work, \citet{Huang2023Improving} show that code-switching improves the retrieval performance on low-resource languages, however, their focus lies on CLIR with English documents. 
To the best of our knowledge, we are the first to systematically investigate (1)~artificial code-switching to train CEs and (2)~the interaction between MoIR, CLIR and MLIR. 

\textbf{Our contributions} are as follows: (i) We show that training on artificially code-switched data improves zero-shot cross-lingual and multilingual rankers. (ii) We demonstrate its robustness towards the ratio of code-switched tokens and effectiveness in generalizing to unseen languages. (iii) We release our code and resources.\footnote{\href{https://github.com/MaiNLP/CodeSwitchCLIR}{https://github.com/MaiNLP/CodeSwitchCLIR}}

\section{Methodology}
\label{s:methodology}

\paragraph{Reranking with Cross-Encoders.}

We follow the standard cross-encoder reranking approach (CE) proposed by \citet{nogueira2019passage}, which formulates relevance prediction as a sequence pair (query-document pair) classification task. CEs are composed of an encoder model and a relevance prediction model. The encoder is a pre-trained language model \citep{devlin-etal-2019-bert} that transforms the concatenated input \texttt{[CLS]~Q~[SEP]~D~[SEP]} into a joint query-document feature representation, from which the classification head predicts relevance. Finally, documents are reranked according to their predicted relevance. 
We argue that fine-tuning CEs on monolingual data biases the encoder towards encoding features that are only useful when the target setup is MoIR. To mitigate this bias, we propose to perturb the training data with code-switching, as described next.

\paragraph{Artificial Code-Switching.}
While previous work has studied code-switching (CS) as a natural phenomenon where speakers borrow words from other languages (e.g.\ anglicism) \citep{ganguly-2016-retrievability,wang-etao-2018-naturalcs}, we here refer to code-switching as a method to \textit{artificially} modify monolingual training data. In the following we assume availability of English (EN--EN) training data. The goal is to improve the zero-shot transfer of ranking models into cross-lingual language pairs X--Y by training on code-switched data EN$_{\text{X}}$--EN$_\text{Y}$ instead, which we obtain by exploiting bilingual lexicons similar to \citet{tan-joty-2021-code}. We now describe two CS approaches based on lexicons: one derived from word embeddings and one from Wikipedia page titles (cf. Appendix \ref{subsec:appendix-examples} for examples). %

\paragraph{Code-Switching with Word Embeddings.}
We rely on bilingual dictionaries $\mathcal{D}$ induced from cross-lingual word embeddings \citep{mikolov2013exploiting,heyman-etal-2017-bilingual} and compute for each EN term its nearest (cosine) cross-lingual neighbor. In order to generate ${\text{EN}_\text{X}\text{--}\text{EN}_\text{Y}}$ we then use $\mathcal{D}_{\text{EN}\veryshortarrow \text{X}}$ and $\mathcal{D}_{\text{EN}\veryshortarrow \text{Y}}$ to code-switch query and document terms from EN into the languages X and Y, each with probability $p$. This approach, dubbed Bilingual CS (\textbf{\texttt{BL-CS}}), allows a ranker to learn inter-lingual semantics between EN, X and Y. In our second approach, Multilingual CS (\textbf{\texttt{ML-CS}}), we additionally sample for each term a different target language into which it gets translated; we refer to the pool of available languages as seen languages.

\paragraph{Code-Switching with Wikipedia Titles.} 
Our third approach, \textbf{\texttt{Wiki-CS}}, follows \citep{lan-etal-2020-empirical, fetahu2021gazetteer} and uses bilingual lexicons derived from parallel Wikipedia page titles obtained from inter-language links. We first extract word $n$-grams from queries and documents with different sliding window of sizes $n \in \{1,2,3\}$. Longer $n$-gram are favored over shorter ones in order to account for multi-term expressions, which are commonly observed in named entities. In \texttt{Wiki CS} we create a single multilingual dataset where queries and documents from different training instances are code-switched into different languages.

\section{Experimental Setup}
\label{s:experimental}

\begin{table*}[ht!]
\small
    \centering
    \begin{tabular}{l c c c c c c c c}
    \toprule
 & EN--EN & DE--DE & RU--RU & AR--AR & NL--NL & IT--IT & AVG & $\Delta_{\text{ZS}}$ \\ \midrule
\texttt{Zero-shot} & 35.0 & 25.9 & 23.8 & \textbf{23.9} & 27.2 & 26.9 & 25.5 &- \\
\texttt{Fine-tuning} & 35.0 & 30.3* & 28.5* & 27.2* & 30.8* & 30.9* & 29.5 & +4.0 \\ \cdashline{1-9}[.4pt/1pt]\noalign{\vskip 0.5ex}
\texttt{Zero-shot}$_{\text{Translate Test}}$ & - & 22.5* & 18.2* & 17.7* & 24.7* & 23.3* & 21.3 & -4.2 \\
$\texttt{ML-CS}_{\text{Translate Test}}$ & - & 22.8* & 18.6* & 17.7* & 24.7* & 24.5* & 21.7 & -3.8 \\ \cdashline{1-9}[.4pt/1pt]\noalign{\vskip 0.5ex}
\texttt{BL-CS} & - & \textbf{26.0} & \textbf{25.5} & 23.0 & \textbf{27.5} & \textbf{27.2} & \textbf{25.8} & \textbf{+0.3} \\
\texttt{ML-CS} & 34.0 & 25.9 & 24.7 & 21.3 & 27.2 & 26.9 & 25.2 & -0.3 \\
\texttt{Wiki-CS} & 33.8* & 25.6 & 24.1 & 20.5* & 27.0 & 25.5* & 24.5 & -1.0 \\
\bottomrule
    \end{tabular}
    \caption{MoIR: Monolingual results on mMARCO languages and averaged over all languages (excluding EN--EN) in terms of MRR@10. \textbf{Bold}: Best zero-shot performance for each language. $\Delta_{\text{ZS}}$: Absolute difference to \texttt{Zero-shot}. Results significantly different from \texttt{Zero-shot} are marked with * (paired t-test, Bonferroni correction, $p<0.05$).}    
    \label{tab:moir_results}
\end{table*}

\setlength{\tabcolsep}{3.8pt}
\begin{table*}[ht!]
\small
    \centering
    \begin{tabular}{l c c c c c c c c c c c }
    \toprule
     & EN--DE & EN--IT & EN--AR & EN--RU & DE--IT & DE--NL & DE--RU & AR--IT & AR--RU & AVG & $\Delta_{\text{ZS}}$ \\ \midrule
\texttt{Zero-shot} & 24.0 & 23.0 & 14.0 & 18.3 & 15.0 & 19.7 & 12.9 & 7.7 & 7.1 & 15.7 & - \\
\texttt{Fine-tuning} & 29.7* & 30.5* & 26.5* & 28.0* & 26.9* & 27.9* & 25.5* & 23.9* & 22.7* & 26.8 & +11.1 \\
\cdashline{1-12}[.4pt/1pt]\noalign{\vskip 0.5ex}
\texttt{Zero-shot}$_{\text{Translate Test}}$ & 22.8 & 23.2 & 16.4 & 17.0 & 15.8 & 17.5 & 11.8 & 9.8 & 8.7 & 15.9 & +0.2 \\
$\texttt{ML-CS}_{\text{Translate Test}}$ & 24.9 & 24.6 & 17.9* & 19.5 & 17.6 & 19.3* & 14.3 & 12.2* & 10.6* & 17.9 & +2.2 \\
\cdashline{1-12}[.4pt/1pt]\noalign{\vskip 0.5ex}
\texttt{BL-CS} & \textbf{26.9*} & \textbf{27.3*} & 19.3* & 22.8* & \textbf{20.4*} & \textbf{22.8*} & 17.8* & \textbf{15.6*} & 14.1* & \textbf{20.8} & \textbf{+5.1} \\
\texttt{ML-CS} & 26.5* & 26.4* & 18.1* & 22.1* & 19.8* & \textbf{22.8*} & 17.8* & 15.3* & 14.2* & 20.3 & +4.6 \\
\texttt{Wiki-CS} & 26.2* & 26.4* & \textbf{19.4*} & \textbf{22.9*} & 19.4* & 22.4*  & \textbf{18.3*} & 14.4* & 14.1* & 20.4 & +4.7 \\ \bottomrule
    \end{tabular}
    \caption{CLIR: Cross-lingual results on mMARCO in terms of MRR@10.}
    \label{tab:clir_results}
\end{table*}

\begin{table*}[ht!]
\small
    \centering
    \begin{tabular}{l ccc cc ccc cc}
    \toprule
& \multicolumn{3}{c}{Seen Languages} & & & \multicolumn{3}{c}{All Languages}  \\
\cmidrule(lr){2-4} \cmidrule(lr){7-9}
& X--EN & EN--X & X--X & $\text{AVG}_{\text{seen}}$ & $\Delta_{\text{seen}}$ & X--EN & EN--X & X--X & $\text{AVG}_{\text{all}}$ & $\Delta_{\text{all}}$ \\ \midrule
\texttt{Zero-shot} & 19.0 & 23.5 & 16.3& 19.6 & - & 16.5 & 20.8 & 12.9 & 16.6 & - \\
\texttt{Fine-tuning} & 24.8* & 26.4* & 21.1* & 24.1 & +4.5 & 26.5* & 26.5* & 21.9* & 25.0 & +8.3 \\ \cdashline{1-11}[.4pt/1pt]\noalign{\vskip 0.5ex}
\texttt{ML-CS} & \textbf{24.2*} & 25.9* & \textbf{21.1*} & \textbf{23.7} & +4.1 & \textbf{21.6*} & 23.2* & 17.0* & 20.6 & +3.9 \\
\texttt{Wiki-CS} & 23.6* & \textbf{26.0*} & 20.6* & 23.4 & +3.8 & 21.3* & \textbf{23.8*} & \textbf{17.1*} & \textbf{20.7} & \textbf{+4.0} \\
\bottomrule
    \end{tabular}
    \caption{MLIR: Multilingual results on mMARCO in terms of MRR@10. Left: Six seen languages for which we used bilingual lexicons to code-switch training data. Right: All fourteen languages included in mMARCO.}
    \label{tab:mlir_results}
    \vspace{-0.1cm}
\end{table*}

\paragraph{Models and Dictionaries.} 
We follow \citet{bonifacio2021mmarco} and initialize rankers with the multilingual encoder mMiniLM provided by \citet{reimers-gurevych-2020-making}. 
We report hyperparameters in Appendix \ref{a:hps}. 
For \texttt{BL-CS} and \texttt{ML-CS}
we use multilingual MUSE embeddings\footnote{\href{https://github.com/facebookresearch/MUSE}{https://github.com/facebookresearch/MUSE}} to induce bilingual lexicons \citep{lample2018word}, which have been aligned with initial seed dictionaries of 5k word translation pairs. We set the translation probability $p=0.5$. %
For \texttt{Wiki-CS}, we use the lexicons provided by the linguatools project.\footnote{\href{https://linguatools.org/tools/corpora/wikipedia-parallel-titles-corpora/}{https://linguatools.org/wikipedia-parallel-titles}} 

\paragraph{Baselines.} To compare whether training on CS'ed data ${\text{EN}_{\text{X}}\text{--EN}_\text{Y}}$ improves the transfer into CLIR setups, we include the zero-shot ranker trained on ${\text{EN--EN}}$ as our main baseline (henceforth, \texttt{Zero-shot}). Our upper-bound reference, dubbed \texttt{Fine-tuning}, refers to ranking models that are directly trained on the target language pair X--Y, i.e.\ no zero-shot transfer. Following \citet{roy-etal-2020-lareqa}, we adopt the \textit{Translate Test} baseline and translate any test data into EN using using our bilingual lexicons induced from word embeddings. On this data we evaluate both the \texttt{Zero-shot} baseline ($\texttt{Zero-shot}_{\text{Translate Test}}$) and our \texttt{ML-CS} model ($\texttt{ML-CS}_{\text{Translate Test}}$).

\paragraph{Datasets and Evaluation.} We use use the publicly available multilingual mMARCO data set~\citep{bonifacio2021mmarco}, which includes fourteen different languages. We group those into six seen languages (EN, DE, RU, AR, NL, IT) and eight unseen languages (HI, ID, IT, JP, PT,  ES, VT, FR) and construct a total of 36 language pairs.\footnote{Due to computational limitations we don't exhaustively evaluate on all possible language pairs.} Out of those, we construct setups where we have documents in different languages (EN--X), queries in different languages (X--EN), and both in different languages (X--X). Specifically, for each document ID (query ID) we sample the content from one of the available languages. For evaluation, we use the official evaluation metric MRR@10.\footnote{We use the implementation provided by the \texttt{ir-measures} package \citep{macavaney2022streamlining}.} All models re-rank the top 1,000 passages provided for the passage re-ranking task. We report all results as averages over three random seeds. 

\section{Results and Discussion}
\label{s:results}

We observe that code-switching improves cross-lingual and multilingual re-ranking, while not impeding monolingual setups, as shown next.

\paragraph{Transfer into MoIR vs.\ CLIR.}
We first quantify the performance drop when transferring models trained on EN--EN to MoIR as opposed to CLIR and MLIR. Comparing \texttt{Zero-shot} results between different settings we find that the average MoIR performance of 25.5 MRR@10 (Table~\ref{tab:moir_results}) is substantially higher than CLIR with 15.7 MRR@10 (Table~\ref{tab:clir_results}) and MLIR with 16.6 MRR@10 (Table~\ref{tab:mlir_results}). The transfer performance greatly varies with the language proximity, in CLIR the drop is larger for setups involving typologically distant languages (AR--IT, AR--RU), to a lesser extent the same observation holds for MoIR (AR--AR, RU--RU). This is consistent with previous findings made in other syntactic and semantic NLP tasks \citep{he-etal-2019-cross,lauscher-etal-2020-zero}. 
The performance gap to \texttt{Fine-tuning} on translated data is much smaller in MoIR (+4 MRR@10) than in CLIR (+11.1 MRR@10) and MLIR (+8.3 MRR@10). Our aim to is close this gap between zero-shot and full fine-tuning in a resource-lean way by training on code-switched queries and documents.

\paragraph{Code-Switching Results.} 
Training on code-switched data consistently outperforms zero-shot models in CLIR and MLIR (Table~\ref{tab:clir_results} and Table~\ref{tab:mlir_results}). In AR--IT and AR--RU we see improvements from 7.7 and 7.1 MRR@10 up to 15.6 and 14.1 MRR@10, rendering our approach particularly effective for distant languages. %
Encouragingly, Table~\ref{tab:moir_results} shows that the differences between both of our CS approaches (\texttt{BL-CS} and \texttt{ML-CS}) versus \texttt{Zero-shot} %
is not statistically significant, showing that gains can be obtained without impairing MoIR performance. Table~\ref{tab:clir_results} shows that specializing one zero-shot model for multiple CLIR language pairs (\texttt{ML-CS}, \texttt{Wiki-CS}) performs almost on par with specializing one model for each language pair (\texttt{BL-CS}). The results of \texttt{Wiki-CS} are slightly worse in MoIR and on par with \texttt{ML-CS} on MLIR and CLIR.

\paragraph{Translate Test vs.\ Code-Switch Train.}
In MoIR (Table \ref{tab:moir_results}) both $\texttt{Zero-shot}_{\text{Translate Test}}$ and $\texttt{ML-CS}_{\text{Translate Test}}$ underperform compared to other approaches. This shows that zero-shot rankers work better on clean monolingual data in the target language than noisy monolingual data in English. In CLIR, where \textit{Translate Test} bridges the language gap between X and Y, we observe slight improvements of +0.2 and +2.2 MRR@10 (Table \ref{tab:clir_results}). However, in both MoIR and CLIR \textit{Translate Test} consistently falls behind code-switching at training time.

\paragraph{Multilingual Retrieval and Unseen Languages.}
Here we compare how code-switching fares against \texttt{Zero-shot} on languages to which neither model has been exposed to at training time.  Table~\ref{tab:mlir_results} shows the gains remain virtually unchanged when moving from six seen (+4.1 MRR@10 / +3.8 MRR@10) to fourteen languages including eight unseen languages (+3.9 MRR@10 / +4.0 MRR@10). Results in Appendix~\ref{a:unseen_langs} confirm that this holds for unseen languages on the query, document and both sides, suggesting that the best pivot language for zero-shot transfer \citep{turc2021revisiting} may not be monolingual but a code-switched language. %

\begin{figure}[t!]
    \centering
    \includegraphics[width=0.95\linewidth]{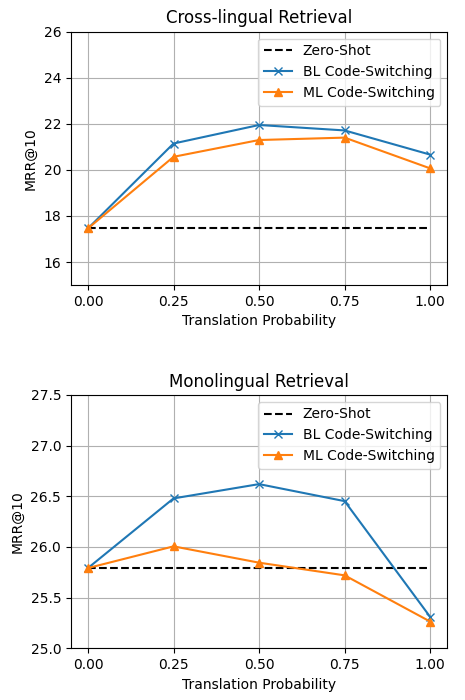}
    \caption{Retrieval performance for different translation probabilities, averaged across all language pairs.}
    \label{fig:ablation}
\end{figure}

\paragraph{Ablation: Translation Probability.}
The translation probability $p$ allows us to control the ratio of code-switched tokens to original tokens, with $p=0.0$ we default back to the \texttt{Zero-shot} baseline, with $p=1.0$ we attempt to code-switch every token.~\footnote{Due to out-of-vocabulary tokens the percentage of translated tokens is slightly lower: 23\% for $p=0.25$, 45\% for $p=0.5$, 68\% for $p=0.75$ and 92\% for $p=1.0$. In Wiki CS 90\% of queries and documents contain at least one translated n-gram, leading to 20\% of translated tokens overall.} Figure~\ref{fig:ablation} (top) shows that code-switching a smaller portion of tokens is already beneficial for the zero-shot transfer into CLIR. The gains are robust towards different values for $p$. The best results are achieved with $p=0.5$ and $p=0.75$ for \texttt{BL-CS} and \texttt{ML-CS}, respectively. Figure~\ref{fig:ablation} (bottom) shows that the absolute differences to \texttt{Zero-shot} are much smaller in MoIR. %

\setlength{\tabcolsep}{6pt}
\begin{table}[t!]
    \centering
    \small
    \begin{tabular}{l c c c}
\toprule
& EN--X & X--EN & X--X \\ \midrule
\multicolumn{4}{l}{\textit{No Code Switching} (\texttt{Zero-Shot})} \\ \midrule
No overlap & 12.2 & 11.0 & 7.4 \\ 
Some overlap & 29.7 & 22.4 & 19.9 \\ 
Significant overlap & 44.6 & 36.4 & 45.5 \\ \cdashline{1-4}[.4pt/1pt]\noalign{\vskip 0.5ex}
All queries & 23.5 & 19.0 & 16.3 \\ \midrule
\multicolumn{4}{l}{\textit{Multilingual Code Switching} (\texttt{ML-CS})} \\ \midrule
No overlap & 15.5 {\scriptsize (+3.3)} & 17.8 {\scriptsize (+6.8)} & 13.0 {\scriptsize (+5.6)} \\ 
Some overlap & 31.7 {\scriptsize (+2.0)} & 27.2 {\scriptsize (+4.8)} & 25.3 {\scriptsize (+5.4)} \\  
Significant overlap & 44.7 {\scriptsize (+0.2)} & 37.8 {\scriptsize (+1.4)} & 45.1 {\scriptsize (-0.5)} \\ \cdashline{1-4}[.4pt/1pt]\noalign{\vskip 0.5ex}
All queries & 25.9 {\scriptsize (+2.4)} & 24.2 {\scriptsize (+5.3)} & 21.1 {\scriptsize (+4.8)} \\ \bottomrule
    \end{tabular}
    \caption{MLIR results on seen languages (MRR@10) broken down into queries share no common tokens (no overlap), between one and three tokens (some overlap) and more than three tokens (significant overlap) with their relevant documents.  Gains of \texttt{ML-CS} are shown in brackets. \mbox{EN--X} has 3,116 queries with no overlap, 3,095 with some overlap and 769 with significant overlap. X--EN has 3,147 queries with no overlap, 2,972 with some overlap and 861 with significant overlap. X--X has 3,671 queries with no overlap, 2,502 with some overlap and 807 with significant overlap.}
    \label{tab:monolingual_overfitting}
\end{table}

\paragraph{Monolingual Overfitting.}
Exact matches between query and document keywords is a strong relevance signal in MoIR, but does not transfer well to CLIR and MLIR due to mismatching vocabularies. Training zero-shot rankers on monolingual data biases rankers towards learning features that cannot be exploited at test time. %
Code-Switching reduces this bias by replacing exact matches with translation pairs,\footnote{We analyzed a sample of 1M positive training instances and found a total of 4,409,974 overlapping tokens before and 3,039,750 overlapping tokens after code-switching (\texttt{ML-CS}, $p=0.5$), a reduction rate of \textasciitilde 31\%.} steering model training towards learning interlingual semantics instead. To investigate this, we group queries by their average token overlap with their relevant documents and evaluate each group separately on MLIR.\footnote{We use the model's SentencePiece tokenizer \citep{kudo-richardson-2018-sentencepiece} and ignore the special tokens \texttt{<s>}, \texttt{</s>}, \texttt{<pad>}, \texttt{<unk>} and \texttt{<mask>}.} The results are shown in Table~\ref{tab:monolingual_overfitting}. Unsurprisingly, rankers work best when there is significant overlap between query and document tokens. However, the performance gains resulting from training on code-switched data (\texttt{ML-CS}) are most pronounced for queries with some token overlap (up to +5.4 MRR@10) and no token overlap (up to +6.8 MRR@10). On the other hand, the gains are much lower for queries with more than three overlapping tokens and range from {-0.5} to +1.4~MRR@10. This supports our hypothesis that code-switching indeed regularizes monolingual overfitting.

\section{Conclusion}
\label{s:conclusion}

We propose a simple and effective method to improve zero-shot rankers:\ training on artificially code-switched data. We empirically test our approach on 36 language pairs, spanning monolingual, cross-lingual, and multilingual setups. 
Our method outperforms zero-shot models trained only monolingually and provides a resource-lean alternative to MT for CLIR. In MLIR our approach can match MT performance while relying only on bilingual dictionaries. To the best of our knowledge, this work is the first to propose artificial code-switched training data for cross-lingual and multilingual IR. %

\section*{Limitations}
This paper does not utilize any major linguistic theories of code-switching, such as \cite{belazi1994code,myers1997duelling,poplack2013sometimes}. Our approach to generating code-switched texts replaces words with their synonyms in target languages, looked up in a bilingual lexicon. Furthermore, we do not make any special efforts to resolve word sense or part-of-speech ambiguity. To this end, the resulting sentences may appear implausible and incoherent.

\section*{Acknowledgements}
We thank the members of the MaiNLP research
group as well as the anonymous reviewers for
their feedback on earlier drafts of this paper. This research is in parts supported
by European Research Council (ERC) Consolidator Grant DIALECT 101043235.

\bibliography{anthology,custom}
\bibliographystyle{acl_natbib}

\onecolumn
\appendix

\section{Code-Switching Examples}
\label{subsec:appendix-examples}

\definecolor{russian}{HTML}{D55E00}  
\definecolor{arabic}{HTML}{DC267F}  
\definecolor{german}{HTML}{009E73}  
\definecolor{italian}{HTML}{0072B2}
\definecolor{dutch}{HTML}{5D3A9B}

\begin{table*}[ht!] 
    \centering \doublespacing
    \small
    \begin{tabular}{p{0.11\linewidth} p{0.35\linewidth} p{0.48\linewidth}}
        \toprule
         {\normalsize Approach} & {\normalsize Query} & {\normalsize Document} \\ \midrule
         \texttt{Zero-Shot} & What is an affinity credit card program? & Use your PayPal Plus credit card to deposit funds. If you have a PayPal Plus credit card, you are able to instantly transfer money from it to your account. This is a credit card offered by PayPal for which you must qualify. \\ \cdashline{1-3}
         
         \texttt{Fine-tuning} & \textcolor{german}{Was ist ein Affinity-Kreditkartenprogramm?} & \textcolor{russian}{\foreignlanguage{russian}{Используйте свою кредитную карту}} PayPal Plus \textcolor{russian}{\foreignlanguage{russian}{для внесения средств. Если у вас есть кредитная карта}} PayPal Plus, \textcolor{russian}{\foreignlanguage{russian}{вы можете мгновенно переводить деньги с нее на свой счет. Это кредитная карта, предлагаемая}} PayPal, \textcolor{russian}{\foreignlanguage{russian}{на которую вы должны претендовать.}} \\ \cdashline{1-3}
         
         \texttt{BL-CS} & \textcolor{german}{Denn is einem} affinity credit card programms? & \textcolor{russian}{\foreignlanguage{russian}{Использовать}} your PayPal \textcolor{russian}{\foreignlanguage{russian}{плюс кредита билет попытаться депозиты}} funds. \textcolor{russian}{\foreignlanguage{russian}{если}} you have a PayPal \textcolor{russian}{\foreignlanguage{russian}{плюс}} credit \textcolor{russian}{\foreignlanguage{russian}{билет}}, \textcolor{russian}{\foreignlanguage{russian}{скажите}} are able to instantly \textcolor{russian}{\foreignlanguage{russian}{переход денег}} from it \textcolor{russian}{\foreignlanguage{russian}{попытаться ваши}} account. This is a credit \textcolor{russian}{\foreignlanguage{russian}{билет}} offered by paypal for \textcolor{russian}{\foreignlanguage{russian}{причём}} you \textcolor{russian}{\foreignlanguage{russian}{может}} qualify. \\ \cdashline{1-3}
         
         \texttt{ML-CS} &   \setcode{utf8} What is \textcolor{russian}{\foreignlanguage{russian}{это}} \textcolor{italian}{affinità} credit card program? & Use \textcolor{dutch}{jouw} PayPal Plus credit \textcolor{dutch}{geheugenkaarten} to \textcolor{italian}{depositi} funds. \textcolor{arabic}{\<إذا>} you \textcolor{russian}{\foreignlanguage{russian}{хотя}} \textcolor{german}{ein} \textcolor{arabic}{\<الائتمان>} \textcolor{italian}{aggiunta} credit card, you are \textcolor{russian}{\foreignlanguage{russian}{попытаться}} \textcolor{italian}{quindi} \textcolor{german}{sofort} transfer \textcolor{german}{geld} from \textcolor{italian}{questo} \textcolor{arabic}{\<إلى>} \textcolor{german}{deine} account. \textcolor{russian}{\foreignlanguage{russian}{Это является}} a \textcolor{russian}{\foreignlanguage{russian}{кредита}} card \textcolor{italian}{offerto} by paypal \textcolor{dutch}{voor} which you \textcolor{italian}{devono} \textcolor{arabic}{\<للتأهل>} \\ \cdashline{1-3}

         \texttt{Wiki-CS} & What is an affinity \textcolor{german}{Kreditkarte} program? & Use your PayPal Plus \textcolor{italian}{carta di credito} to deposit funds. If you have a PayPal Plus \textcolor{italian}{carta di credito}, you are able to instantly transfer \textcolor{italian}{denaro} from it to your account. This is a \textcolor{italian}{carta di credito} offered by PayPal for which you \textcolor{italian}{mosto} qualify. \\ \bottomrule

    \end{tabular}
    \caption{Different Code-Switching strategies on a single training instance for the target language pair DE--RU (Query ID: 711253, Document ID: 867890, label: 0). \textbf{\texttt{Zero-shot}:} Train a single zero-shot ranker on the original EN--EN MS MARCO instances \citep{bajaj2016ms}. \textbf{\texttt{Fine-tuning}:} Fine-tune  ranker directly on DE--RU, we use translations (Google Translate) provided by the mMARCO dataset \cite{bonifacio2021mmarco}. \textbf{Bilingual Code-Switching (\texttt{BL-CS})}: Translate randomly selected EN query tokens into DE and randomly selected EN document tokens into RU, each token is translated with probability $p=0.5$; \textbf{Multilingual Code-Switching (\texttt{ML-CS})}: Same as \texttt{BL-CS} but additionally sample for each token its target language uniformly at random. \textbf{\texttt{Wiki-CS}}: Translate $n$-grams extracted with a sliding window. Tokens within a single query/document are code-switched with a single language; across training instances languages are randomly mixed. We use the following ``seen languages'': English, \textcolor{german}{German}, \textcolor{russian}{Russian}, \textcolor{italian}{Italian}, \textcolor{dutch}{Dutch}, \textcolor{arabic}{Arabic}.}
    \label{tab:my_label}
\end{table*}

\clearpage

\section{Results on Unseen Languages}
\label{a:unseen_langs}

\setlength{\tabcolsep}{4.8pt}
\begin{table}[ht!]
    \small
    \centering
    \begin{tabular}{l ccc ccc ccc cc}
    \toprule
    & \multicolumn{2}{c}{Unseen QL} & \multicolumn{3}{c}{Unseen DL} &\multicolumn{4}{c}{Unseen Both} \\
    \cmidrule(lr){2-3} \cmidrule(lr){4-6} \cmidrule(lr){7-10}
    & FR--EN & ID--NL & EN--PT & DE--VT & IT--ZH & ES--FR & FR--PT & ID--VT & PT--ZH & AVG & $\Delta_{\text{ZS}}$ \\ \midrule
Zero-shot & 18.3 & 13.7 & 23.2 & 10.9 & 9.4 & 19.0 & 18.7 & 11.8 & 9.6 & 15.0 & - \\
Fine-tuning & 30.0* & 27.2* & 30.8* & 24.8* & 25.0* & 29.0* & 29.0* & 25.8* & 25.4* & 27.4 & +12.2 \\ \cdashline{1-12}[.4pt/1pt]\noalign{\vskip 0.5ex}
Multilingual CS & \textbf{21.4*} & \textbf{18.3*} & 25.9* & \textbf{15.5*} & 14.8* & \textbf{22.7*} & \textbf{21.9*} & \textbf{16.4*} & 14.7* & \textbf{19.1} & \textbf{+4.1} \\
Wiki CS & 21.0* & 17.2* & \textbf{26.2*} & 15.4* & \textbf{15.0*} & 21.9* & 20.5* & 15.3* & \textbf{14.8*} & 18.6 & +3.4 \\
\bottomrule
    \end{tabular}
    \caption{CLIR results on unseen mMARCO languages in terms of MRR@10. \textbf{Bold}: Best zero-shot model for each language pair. $\Delta_{\text{ZS}}$: Absolute difference to the zero-shot baseline. Results significantly different from the zero-shot baseline are marked with * (paired t-test, Bonferroni correction, $p<0.05$). Results include unseen query languages (QL), unseen document languages (DL) and unseen languages on both sides.}
    \label{tab:clir_unseen_results}
\end{table}

\begin{table}[ht!]
    \small
    \centering
    \begin{tabular}{l c c c c c c cc}
    \toprule
& FR--FR & ID--ID & ES--ES & PT--PT & ZH--ZH & VT--VT & AVG & $\Delta_{\text{ZS}}$ \\ \midrule
Zero-shot & \textbf{27.2} & \textbf{26.8} & \textbf{28.2} & \textbf{27.9} & \textbf{24.8} & 22.8 & 26.3 & - \\ 
Fine-tuning & 30.5* & 30.6* & 31.5* & 31.2* & 29.1* & 28.6* & 30.3 & +4.0\\ \cdashline{1-9}[.4pt/1pt]\noalign{\vskip 0.5ex}
Multilingual CS & 26.4 & 26.7 & 27.6 & 27.3 & 22.3 & \textbf{23.1*} & 25.6 & -0.7\\
Wiki CS & 25.8* & 25.5* & 27.1* & 26.5* & 22.2* & 21.8* & 24.8 & -1.8 \\
\bottomrule
    \end{tabular}
    \caption{MoIR: Monolingual results on unseen mMARCO languages in terms of MRR@10.}
    \label{tab:moir_unseen_results}
\end{table}

\section{Hyperparameters, Datasets and Infrastructure} \label{a:hps}
\begin{table}[htp!]
    \centering
    \begin{tabular}{l r}
    \toprule
    Hyperparameter & Value \\
    \midrule
         Maximum sequence length & 512   \\
         Learning rate & 2e-5 \\
         Training steps & 200,000 \\
         Batch size & 64 \\
         Warm-up steps (linear) & 5,000 \\
         Positive-to-negative ratio & 1:4 \\
         Optimizer & AdamW \citep{loshchilov2018decoupled}\\
         Encoder Model & \texttt{nreimers/mMiniLMv2-L6-H384-distilled-from-XLMR-Large} \\
         Encoder Parameters & 106,993,920 \\
        \bottomrule
    \end{tabular}
    \caption{Hyperparameter values for re-ranking models. Following \citet{reimers-gurevych-2020-making} we extract negative samples from training triplets provided by MS MARCO  \cite{bajaj2016ms}. In the passage re-ranking task we re-rank for 6980 queries 1,000 passages respectively (qrels.dev.small). We construct 36 different language pairs from the mMARCO dataset \cite{bonifacio2021mmarco}.} 
    \label{tab:hps}
\end{table}

\setlength{\tabcolsep}{18.8pt}
\begin{table}[htp!]
    \centering
    \begin{tabular}{l r}
        \toprule 
        Setup & \\ \midrule
        GPU & NVIDIA A100 (80 GB) \\
        Avg. Training Duration (per model) & 13 h \\
        Avg. Test (per language pair) & 2 h \\
        \bottomrule
    \end{tabular}
    \caption{Computational environment. We use \texttt{Huggingface} to train our models \citep{wolf-etal-2020-transformers}, NLTK for tokenization, \texttt{ir-measures} for evaluating MRR@10 \citep{macavaney2022streamlining} and \texttt{SciPy} for significance testing.}
    \label{tab:env}
\end{table}
\clearpage

\section{Bilingual Lexicon Sizes}

\begin{table}[htp!]
    \centering
\begin{tabular}{lrr}
\toprule
Language & MUSE vocabulary & Parallel Wikipedia titles \\
\midrule
Arabic & 132,480  &  432,359 \\
German & 200,000  &    1,113,422 \\
Italian & 200,000  &     999,243 \\
Dutch & 200,000  &     822,563 \\
Russian  & 200,000  &     906,750 \\
\bottomrule
\end{tabular}
    \caption{Size of bilingual lexicons. Two lexicons are used to substitute the words in English with their respective cross-lingual synonyms: (i) multilingual word embeddings provided by MUSE \cite{lample2018word}, (ii) Wikipedia page titles obtained from inter-language links, provided by linguatools project.\footnote{\href{https://linguatools.org/tools/corpora/wikipedia-parallel-titles-corpora/}{linguatools.org/wikipedia-parallel-titles}} The Wikipedia-based lexicons are several times larger that the MUSE vocabulary.}
    \label{tab:wiki_lexicons}
\end{table}

\end{document}